\documentclass[runningheads]{llncs}
\usepackage{graphicx}
\usepackage{amsfonts}
\usepackage{amsmath}
\usepackage{xspace} 
\usepackage{comment}
\usepackage{todonotes}
\usepackage{subcaption}
\usepackage{tabularx}
\usepackage{multirow}
\usepackage{colortbl}
\usepackage{pgfplots}
\begin{document}
\title{Differential Privacy for Anomaly Detection: Analyzing the Trade-off Between Privacy and Explainability}
\setlength{\parskip}{2pt}
\titlerunning{DP for AD: Analyzing the Trade-off Between Privacy and Explainability}
%
\author{Fatima Ezzeddine \thanks{F. Ezzeddine and M. Saad are co-first author} \inst{1,2} \and
Mirna Saad$^{*}$\inst{3} \and
Omran Ayoub\inst{1}\and
Davide Andreoletti \inst{1} \and
Martin Gjoreski \inst{2}\and
Ihab Sbeity \inst{3}\and
Marc Langheinrich \inst{2}\and
Silvia Giordano \inst{1}
}
\authorrunning{F. Ezzeddine, M. Saad et al.}
\institute{University of Applied Sciences and Arts of Southern Switzerland, Switzerland \and
Università della Svizzera italiana, Switzerland \and
Lebanese University, Lebanon \\
\email{Corresponding author: fatima.ezzeddine@usi.ch}
}
\maketitle              
\begin{abstract}
Anomaly detection (AD), also referred to as outlier detection, is a statistical process aimed at identifying observations within a dataset that significantly deviate from the expected pattern of the majority of the data. Such a process finds wide application in various fields, such as finance and healthcare. While the primary objective of AD is to yield high detection accuracy, the requirements of explainability and privacy are also paramount. The first ensures the transparency of the AD process, while the second guarantees that no sensitive information is leaked to untrusted parties. In this work, we exploit the trade-off of applying Explainable AI (XAI) through SHapley Additive exPlanations (SHAP) and differential privacy (DP).
We perform AD with different models and on various datasets, and we thoroughly evaluate the cost of privacy in terms of decreased accuracy and explainability. Our results show that the enforcement of privacy through DP has a significant impact on detection accuracy and explainability, which depends on both the dataset and the considered AD model. We further show that the visual interpretation of explanations is also influenced by the choice of the AD algorithm.

\keywords{Explainable AI  \and Differential Privacy \and Anomaly Detection.}
\end{abstract}
\section{Introduction}
Within the realm of data-driven decision-making, anomalies, which are data points exhibiting statistically significant deviations from expected patterns, have multifaceted impact. Anomalies are indicative of errors or inconsistencies within the data, but they also hold the potential to reveal novel or critical situations. Deviations are subsequently flagged for further investigation as they can be highly informative, often signaling underlying issues or emerging trends. For instance, in the context of cyber-security, an anomaly might indicate a security breach or an attempted attack. In healthcare, it could highlight rare illnesses. The timely identification of anomalies is crucial for maintaining security, efficiency, and safety in various fields, such as cyber-security, healthcare, network monitoring, and transportation \cite{ahmed2016survey,jiang2021applications}. Therefore, developing highly effective Anomaly Detection (AD) systems is of paramount importance, as they represent a critical tool to address the challenge of detecting these anomalies \cite{chandola2009anomaly}. By leveraging diverse statistical and machine learning (ML) techniques, AD establishes a statistical baseline for normal behavior within a dataset.

Recently, a main requirement of AD systems emerged, in addition to that of high predictive performance, which is the need of providing stakeholders with relevant information on why and how a specific data point is considered an anomaly, such as to enhance the transparency of AD systems, with the final aim of fostering trust in these systems \cite{guidotti2018survey,alharbi2022explainable,roshan2022using}. 
In addition to that, privacy guarantees are also essential for AD in scenarios as data owners may delegate the AD task to a third party in possession of the technical expertise needed for effective AD. This third-party data access introduces privacy concerns, especially when dealing with sensitive data such as healthcare information that includes confidential patient medical histories \cite{yang2019tradeoff,mayer2020privacy,degue2021differentially}. In this context, we are confronted with the challenge of ensuring both transparency and privacy in AD systems. To ensure the privacy of data and the transparency of decisions, differential privacy (DP) \cite{dwork2006differential} and explainable artificial intelligence (XAI) have been the main methods attracting the attention of the research community, respectively.
A conflict however arises between ensuring transparency (through XAI) and privacy (through DP) due to their opposing goals. XAI aims to provide insights into model behavior for transparency, while privacy-preserving solutions obscure data to prevent data leakage. Specifically, XAI techniques are proposed to demystify the inner workings of complex ML models and AD through different types of explanations such as, e.g., feature importance, which scores the contribution and impact of each feature on the model's output, permitting data owners to identify which features in the data were most influential in identifying an anomaly. DP, on the contrary, works by injecting calibrated noise into the data before it is released, introducing a quantifiable privacy guarantee, and allowing data owners to control the level of information revealed about individual data points. 

The intersection of XAI with privacy-preserving techniques presents complex and nuanced challenges. As privacy concerns arise, the implementation of privacy-preserving mechanisms often involves obfuscating sensitive information, which may not only impact the performance and utility of AD models but also their explainability. Therefore, there exists a critical need to precisely quantify how privacy-preserving techniques such as DP affect the explainability of any ML or AD systems \cite{linardatos2020explainable}. In this paper, to investigate the complex relationship among DP, AD, and XAI. As a XAI framework, we rely on SHapley Additive exPlanations (SHAP) \cite{lundberg2017unified}. To investigate this interplay in the context of AD, we formulate two research questions (RQs) and address them as follows:
\begin{enumerate}
    \item \emph{RQ.1 To what extent does increasing DP noise affect the fidelity and stability of SHAP values in AD?} We explore XAI in DP-AD, specifically, we investigate SHAP with a focus on understanding how the SHAP values of AD models are affected by varying levels of DP noise on AD. By extensively analyzing quantitatively and qualitatively the output of SHAP, we shed light on the potential trade-offs between privacy protection and model explainability. 

    \item \emph{RQ2. To what extent does the application of DP impact the performance and explainability of AD algorithms designed for specific anomaly types (local vs. global)?}
    Leveraging the inherent specialization of AD algorithms towards distinct anomaly types (local vs. global), this research delves into the potential for DP and its impact on their performance and explainability under varying privacy constraints.
\end{enumerate}

The paper is organized as follows. Section \ref{RelatedWork} discusses related work. Section \ref{BackroundSection} presents background on AD and DP. Section \ref{ProblemFormulation} details the problem and objectives. Section \ref{Evaluation} presents the experimental setup and evaluation settings. Section \ref{Results} showcases the obtained results, and analyzes their implications, offering valuable insights for practitioners navigating the trade-off between privacy and explainability in AD models. Section \ref{Conclusion} concludes the paper.

\section{Related Work}\label{RelatedWork}

In this section, we first discuss studies that have applied either privacy-preserving or explainability techniques to AD and then studies that have investigated the impact of privacy on the model's explainability.

The related works highlight the growing tension between privacy and explainability. While research has explored privacy-preserving techniques and explainable models in many contexts, their intersection with AD remains largely unaddressed. This is particularly crucial because AD often deals with sensitive data and requires different considerations compared to traditional classification and regression tasks. In this paper, we analyze this intersection to reveal insights into the trade-offs involved, guiding the development of future robust, explainable, and privacy-preserving AD methods that empower informed decision-making.

\subsection{Privacy-Preserving Anomaly Detection}
Anomaly detection methods play a crucial role in diverse sectors such as finance, healthcare, transportation, and smart grids ~\cite{degue2021differentially,ahmed2016surveyfinancial,hassan2020differential,jiang2021applications}. Most existing methods rely primarily on unsupervised ML techniques (e.g., ~\cite{zong2018deep,bergmann2019mvtec,leung2005unsupervised,munir2018deepant,chandola2009anomaly}), with some supervised approaches as well (e.g., ~\cite{jia2019anomaly}). Numerous AD algorithms have been proposed in the literature, and each of them comes with strengths and weaknesses and is suitable for specific contexts and data types. For example, Isolation Forest (iForest) and Local Outlier Factor (LOF) are well suited for tabular data, Deep Learning-based AD with auto-encoders is suitable for images and time series data \cite{chandola2009anomaly,liu2008isolation,chen2018autoencoder}. Several recent studies on AD have proposed employing privacy-preserving techniques to address the critical challenge of balancing effective AD with individual data protection. These techniques mitigate public concern over data sharing, reduce the risk of re-identification from anonymized data, facilitate compliance with data privacy regulations like the General Data Protection Regulation (GDPR), and foster collaboration between organizations, leading to more comprehensive and effective AD across various sectors \cite{degue2021differentially,ma2022privacy}. To achieve AD while safeguarding data privacy, several methods have been employed such as DP \cite{du2019robust,giraldo2020adversarial,angelini2019privacy}, homomorphic encryption and training on encrypted data \cite{mehnaz2020privacy,alabdulatif2019privacy,sridharan2018wadac}, Secure Multi-Party Computation \cite{zhang2012privacy}, and even novel, custom-designed approaches \cite{lyu2016improved}, such as generating synthetic data \cite{mayer2020privacy}, or approaches for specific cases such as body movement and power systems \cite{angelini2019privacy,keshk2019integrated}.

Specifically, among the various techniques, several studies have explored the integration of AD algorithms and DP, e.g., \cite{du2019robust,okada2015differentially}. Notably, Du et al. \cite{du2019robust} show how DP can improve the utility of AD and novelty detection, with a focus on detecting poisoning samples in backdoor attacks. Jiang et. al \cite{jiang2021applications} propose a privacy-preserving social network model that utilizes restricted local DP to sanitize user information collection. Moreover, Chukkapalli et al. in \cite{chukkapalli2021privacy} propose an approach for privacy-preserving AD in smart farming by adding noise to individual farm data. Giraldo et. al \cite{giraldo2020adversarial} examine how AD can be combined with DP to provide robust privacy and security for individuals. In addition to Degue et al. \cite{degue2021differentially} DP is employed with AD in correlated data to analyze the trade-off between privacy level and detection accuracy of multivariate Gaussian signals. 

Other studies have utilized homomorphic encryption and performed training of AD models with encrypted data \cite{alabdulatif2019privacy,guo2019efficient,zhang2012privacy}. For instance, Alabdulatif et al. \cite{alabdulatif2019privacy} focus on cloud-based models and propose an AD model that preserves data privacy with reliance on ciphertext, while, Mehnaz et al. \cite{mehnaz2020privacy} present a framework for efficient AD on real-time-series encrypted data. Guo et al. \cite{guo2019efficient} propose an AD scheme for encrypted video bitstreams with format-compliant encryption. Zhang et al. \cite{zhang2012privacy} propose a semi-centralized privacy-preserving secure multiparty computation protocol for the Principal Component Analysis-based AD. Other approaches have been proposed based on synthesizing data and generating samples, such as \cite{mayer2020privacy}. Mayer et al. \cite{mayer2020privacy} analyze many approaches for creating synthetic data and the utility of the created datasets for AD in supervised, semi-supervised, and unsupervised settings. 
Prioritizing privacy has been the main focus of these recent AD studies. However, explainability in AD is also emerging as a crucial area of research, which we will explore in the following subsection.

\subsection{Explainable Anomaly Detection}
Yuan et al. \cite{yuan2022trustworthy} discuss crucial challenges of AD such as trustworthiness, explainability, and robustness. In this context, several studies explore methods for extracting valuable insights for anomalies explanations \cite{alharbi2022explainable} and as detailed in \cite{panjei2022survey}. For instance, Panjei et al. \cite{panjei2022survey} categorize various types of explanations and analyze existing techniques for interpreting anomalies, paving the way for more meaningful AD analysis. Roshan et al. \cite{roshan2021utilizing} demonstrate the use of XAI to explain the results of a deep learning autoencoder AD model. Similarly, Roshan et al. \cite{roshan2022using} leverages the Kernel SHAP method within XAI to identify and explain network anomalies. Moreover, Ravi et al. \cite{ravi2021general} explore the feasibility and compare the performance of several state-of-the-art XAI frameworks on Convolutional Autoencoders for building reliable and trustworthy AD systems in the visual domain. Finally, Tritscher et al. \cite{tritscher2023feature} highlight the growing interest in categorizing and analyzing these explainable methods based on their access to training data and the specific AD model.

These works have demonstrably investigated interpretability and leveraged XAI techniques for AD, but without considering privacy concerns.
\subsection{Impact of Privacy on Explainability}
Recent years have seen a surge in studies investigating the interplay of privacy with XAI
\cite{bozorgpanah2022privacy,naidu2021differential,ezzeddine2023vertical} as both interpretability and privacy represent a requirement for deploying ML models and datasets. Some studies investigate the inherent trade-off between these concepts, while others propose novel methods to generate privacy-preserving explanations \cite{montenegro2021privacy,veugen2022privacy,jetchev2023xorshap}.

Bozorgpanah et al. \cite{bozorgpanah2022privacy} investigate the impact of various privacy-preserving techniques such as masking, and DP noise addition on the effectiveness of regression-based explainability methods utilizing Shapley values and show different behaviors in Shapley for different models by computing correlation metrics.
Also, the authors in \cite{naidu2021differential} show the impact of DP on the interpretability of Deep Neural Networks particularly in medical imaging application classification, and show significant visual differences in explanation with DP. Another study\cite{de2023xaiprivacy} investigates the use of example-based explainability models for retinal image analysis. The authors propose leveraging Generative Adversarial Networks (GANs) to generate synthetic examples that provide explanations for model predictions while preserving the privacy of the original retinal images. Nori et al. \cite{nori2021accuracy}, a method for adding DP to Explainable Boosting Machines enables the training of interpretable classification and regression models with state-of-the-art accuracy while preserving privacy. Harder et al. \cite{harder2020interpretable} addresses the challenge of balancing interpretability and privacy in ML models by proposing a novel approach using simple models with locally linear maps to approximate complex models. This method achieves high classification accuracy while providing differentially private explanations for the classifications. Montenegro et al. \cite{montenegro2021privacy} propose privacy-preserving GANs for privatizing case-based explanations in classification tasks. This GAN incorporates a counterfactual module, enabling the generation of both factual and counterfactual explanations while safeguarding privacy. Moreover, Jetchev et al. \cite{jetchev2023xorshap} introduces a novel, privacy-preserving algorithm for calculating Shapley values on decision tree ensembles within a secure multi-party computation framework, ensuring data privacy. Veugen et al. \cite{veugen2022privacy} proposes to combine privacy-preserving technologies with state-of-the-art XAI, allowing the generation of privacy-friendly explanations by leveraging local foil trees.

\section{Background}\label{BackroundSection}
This section introduces the concept of AD and the theoretical background relative to the AD algorithms considered in this work, namely LOF and iForest.
\subsection{Anomaly Detection Algorithms}
The term anomalies refers to data points that deviate significantly from the expected patterns or behaviors observed within a dataset. This deviation can indicate various underlying factors, ranging from irregularities to errors in data collection or processing. AD refers to the process of identifying these anomalies~\cite {chandola2009anomaly}. In many applications, practitioners are particularly interested in finding data points that deviate from their immediate neighbors locally (local anomalies) or globally, referring to data points that deviate significantly from the overall distribution of the data set rather than just its immediate neighbors (global anomalies). To address both local and global anomalies, our study incorporates LOF \cite{breunig2000lof} (for local AD), and iForest \cite{liu2008isolation} (for global AD), which are two unsupervised models that proved scalable and efficient \cite{muruti2018survey}.

\subsubsection{Isolation forest} The core principle behind iForest \cite{liu2008isolation} lies in constructing decision trees using randomly sampled data to isolate outliers.
Given the intrinsic sparsity of anomalies relative to normal data points, their isolation pathways within the iForest exhibit demonstrably shorter lengths, therefore, anomalies are isolated faster within the constructed trees. In other words, the fewer branches that need to be traversed in the tree to isolate a data point (indicating a shorter path), the higher the likelihood that it is an anomaly. 

To build a tree, it starts by randomly selecting a feature and a random split value between its minimum and maximum. It then partitions the data into two sets based on the chosen split value. This process is recursively repeated, building out the decision tree structure. A pre-defined maximum depth can be set for each tree to ensure consistency and avoid excessively deep trees. After that, an anomaly score is computed as the average path length of a data point across all the trees in the forest. Lower scores indicate a higher likelihood of being an anomaly, since deeper paths, indicating more effort to isolate, suggest higher normality, while shallower paths, signifying easier isolation, point toward potential anomalies. Since iForest focuses on random partitioning, faster isolation of anomalies, and independence from local density distribution, it is a strong choice for detecting global anomalies that deviate significantly from the overall data distribution.

\vspace{2mm}
\subsubsection{Local outlier factor} LOF \cite{breunig2000lof} algorithm identifies anomalies by comparing the local density of a data point with the density of its k-nearest neighbors. Local density comparison assesses how much an individual point deviates from its surrounding environment. LOF first identifies the k-nearest neighbors for each data point. Then, it calculates a local reachability density (LRD) that estimates how dense the area surrounding a particular data point is compared to its neighbors. Then, LOF calculates a score for each data point by comparing its LRD to the average LRD of its k-nearest neighbors.  Points with a significantly lower LRD compared to their neighbors are considered potential outliers, and higher LOF scores indicate higher local density and normality, while lower scores suggest potential anomalies, deviating significantly from their surrounding data points. LOF focuses on identifying local anomalies and excels at this task by considering local density measures of points. This approach makes LOF ideal for identifying local anomalies that deviate from the norm within their specific local area.

\subsection{Differential Privacy}
DP offers a mathematical framework to ensure individual privacy in data analysis. It achieves this by injecting calibrated noise into various stages of the process, including the input data itself ~\cite{dwork2006differential}, the output of ML models, or even the model weights or internal parameters \cite{abadi2016deep}. DP allows for extracting valuable statistical insights from datasets while demonstrably protecting the privacy of any single record within them. In essence, DP ensures that the overall statistical properties of the dataset remain preserved irrespective of the presence or absence of any specific individual data point in the training set.

A mechanism is defined as any mathematical computation that applies to and interacts with the data. Therefore, if the likelihood of any given result is nearly equal for two datasets that differ by just one record, then the mechanism guarantees DP. The degree of privacy is governed by a parameter known as $\varepsilon$, which dictates how closely the outputs of a DP mechanism resemble each other when applied to two neighboring databases (i.e., datasets that are identical except for the presence or absence of a single individual's data). A smaller $\varepsilon$ offers stronger privacy protection. In Def. \ref{def:differential-privacy}, we detail the DP inequality.
\begin{definition}[Differential Privacy]
A randomized algorithm $M$ with domain $\mathbb{N}^2$ is $(\varepsilon, \delta)$-differentially private if for all $S \subseteq \text{Range}(M)$ and for all $x, y \in \mathbb{N}^2$ such that $\lVert x - y\rVert_1 \leq 1$:
\[
\Pr[M(x) = S] \leq e^{\varepsilon} \cdot \Pr[M(y) \in S] + \delta,
\]
\label{def:differential-privacy}
\end{definition}\vspace{-20pt}
For any subset \(S\) and neighboring datasets \(x, y\), the probabilities of \(M\) on \(x\) are less than \(e^\varepsilon\) times the probabilities on \(y\), plus \(\delta\). This indicates that a randomized algorithm \(M\) is \((\varepsilon, \delta)\)-differentially private.
One common method for achieving DP is by adding Laplace noise to query responses \cite{zhang2021privsyn}. Let $f$ be a function representing the AD algorithm, and $\epsilon$ be the privacy parameter. The Laplace mechanism adds noise according to the formula:
\begin{equation}
    \hat{f}(D) = f(D) + \text{Lap}\left(\frac{\Delta f}{\epsilon}\right)
\end{equation}
Where $\hat{f}(D)$ is the DP query result on dataset $D$, $f(D)$ is the true query result, $\Delta f$ is the sensitivity of the function $f$, and $\text{Lap}(\lambda)$ represents Laplace noise with scale parameter $\lambda$. An alternative approach is to add Gaussian noise to query responses, which adds Gaussian noise instead of Laplacian noise.

\section{Problem Formulation and Approach}\label{ProblemFormulation}
We aim to investigate the effectiveness of AD techniques, specifically iForest and LOF, in identifying two types of anomalies within a dataset: local and global outliers (\emph{RQ2}). This investigation is conducted within a privacy-preserving setup, where we employ DP techniques. Crucially, we also aim to analyze how the SHAP explanations of the AD models change as a result of employing DP. Estimating these explainability changes will allow us to assess the impact of the privacy-preserving mechanisms on the reasoning behind the identified anomalies. To achieve our goals, we will employ Privacy-Preserving AD with DP on the training data level. This means that we will introduce calibrated noise directly to the training data itself before applying the AD algorithms. Following the noise injection into the data, we employ SHAP to analyze how feature contributes to anomaly scores change under different $\varepsilon$-DP guarantees (\emph{RQ1}).
\begin{figure}[h]
    \centering
        \includegraphics[width=\linewidth,height=7cm]{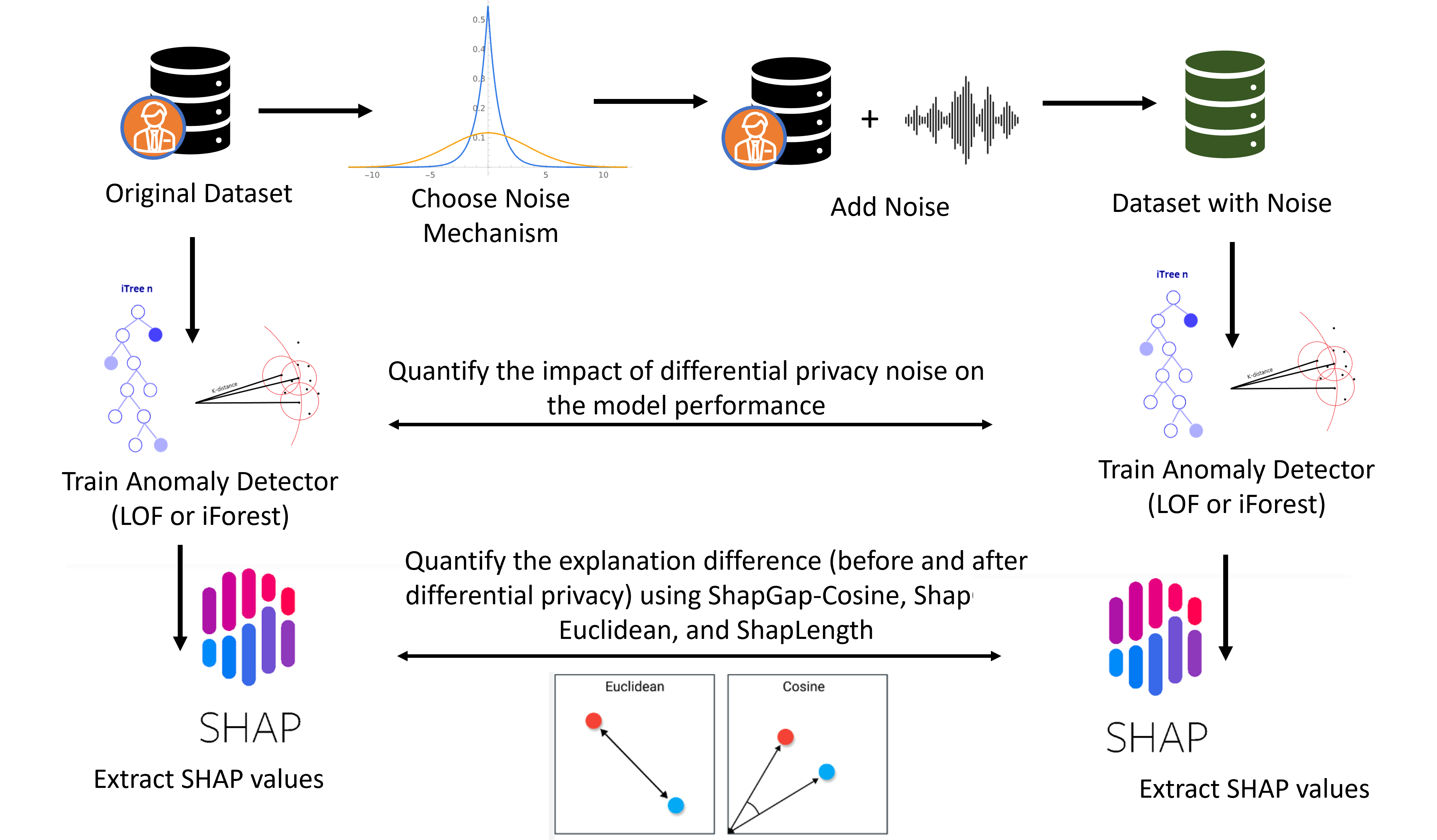}
        \caption{Overall scheme of the experimental setup}
        \label{OverallSetup}
\end{figure}
Figure~\ref{OverallSetup} summarizes the steps followed for the conducted analysis. Let $D$ be a dataset $\mathcal{D}$: $\mathcal{X}$ $\in$ $\mathbb{R}^d$ $\in$ $\mathbb{R}$. The AD algorithm is trained on $D$ in an unsupervised manner to estimate the anomaly score $y_{i}$ for each datapoint $x_{i}$ $\in$ $\mathcal{X}$. 
To ensure a robust level of privacy protection measured by $\varepsilon$-differential privacy ($\varepsilon$-DP), we aim for small values of $\varepsilon$ to minimize the influence of individual records on the overall performance. We introduce noise into the data using either a Laplacian or Gaussian distribution, with varying $\varepsilon$ values (0.01, 0.1, 1, and 5). Subsequently, we retrain the AD models on the data augmented with noise. This process enables us to assess the impact of DP on both the performance and explainability of the models. Specifically, we compare the AD performance of models trained on noisy and original data using diverse metrics, focusing on understanding how DP affects the features driving AD with SHAP. Moreover, we explore the trade-off between privacy and explainability by employing SHAP.

\section{Experimental Setting}\label{Evaluation}
This section describes the settings of our experiments, including the the datasets, the procedures to train and evaluate the AD models, and the metrics used to quantify change in explainability. 

\subsection{Datasets Description}
We consider three datasets that are widely used for AD:
\begin{enumerate}
\item Mammography~\cite{woods1993comparative}: The mammography dataset corresponds to radiological scans to diagnose breast cancer. It consists of 6 features and 11183 records, of which 10923 are non-anomalous and 260 are anomalous.
\item Thyroid dataset~\cite{pang2019deep}: The thyroid dataset corresponds to thyroid diseases. It comprises 21 features and contains 7,200 records, of which 6,666 are non-anomalous and 534 are anomalous.
\item Campaign (bank) dataset~\cite{pang2019deep} includes banking information about individuals. It has 62 features and contains 41188 records, of which 36548 are non-anomalous and 4640 are anomalous.
\end{enumerate}

\subsection{Anomaly Detection Model training}

We perform hyperparameter tuning of the AD models using a Grid search with cross-validation (for a separate subset of the data). The hyperparameters that undergo an optimal selection are  max\_features, n\_estimators for iForest, and n\_neighbors for LOF. It is important to emphasize that all the datasets are labeled. However, as iForest and LOF are unsupervised AD algorithms, we use the labels only for evaluating their performance, as illustrated in Figure \ref{OverallSetup}.

\subsection{Evaluation metrics}\label{EvaluationSec}
\subsubsection{AD Performance Metrics}
We evaluate the performance of the AD model in terms of predictive performance and output consistency. As for the former, we use precision and AUC as metrics. Precision measures the proportion of true positives among all predicted positives, while AUC evaluates the model's ability to distinguish between positive and negative classes. As for the latter, we compute the fidelity score, which measures the degree of agreement between the predictions of two different models on the same set of inputs (i.e., how closely the outputs of one model mirror those of another). In our case, the fidelity score quantifies the agreement between the AD model's outputs before and after applying DP (as shown in Figure \ref{OverallSetup}).

\subsubsection{Quantitative analysis of SHAP difference metrics}

To quantify the SHAP value changes in scenarios with and without DP, we use three recently-proposed metrics, ShapGAP-Euclidean and ShapGAP-Cosine distances \cite{mariotti2023beyond}, and ShapLength \cite{mariotti2022measuring}. While initially proposed for comparing SHAP values between black-box models and their corresponding surrogate white-box models, we leverage these metrics for a different purpose. Specifically, we compute the \emph{ShapGAP} and \emph{ShapLength} between SHAP values of data points before and after adding DP noise, with the final aim of quantifying the impact to assess how DP alters the explainability of AD models. More details about these metrics are provided in the following:

\begin{enumerate}
    \item \textbf{ShapGAP-Euclidean Distance \cite{mariotti2023beyond} (Eq. \ref{euclidean})}: ShapGAP-Euclidean provides a magnitude-based measure of the difference between the SHAP values generated for the same data point from two models $S$ (without and with DP) across $n$ data points $x_i$ of a dataset $D$. It is useful for understanding the overall magnitude of changes.
    \begin{equation}
    ShapGAP_{L2} (D) = \frac{1}{n} \sum_{i}^{n} ||S_{\text{without DP}}(x_i) - S_{\text{with DP}}(x_i)||_2
    \label{euclidean}
    \end{equation}

    \item \textbf{ShapGAP-Cosine Distance \cite{mariotti2023beyond} (Eq. \ref{cosine})}: ShapGAP-Cosine computes the magnitude and directional relationship of the difference between two SHAP values of the same point from two models $S$ (without and with DP) across $n$ data points $x_i$ of a dataset $D$. The ShapGAP-Cosine can range between 0 and 2, where higher values indicate higher dissimilarity. It is useful for capturing how similar the directions are, even if the magnitudes differ.
    \begin{equation}
    ShapGAP_{Cos} (D) = \frac{1}{n} \sum_{i}^{n} (1 - \frac{S_{\text{without DP}}(x_i) \cdot S_{\text{with DP}}(x_i)} {||S_{\text{without DP}}(x_i)||_2 ||S_{\text{with DP}}(x_i)||_2} )
    \label{cosine}
    \end{equation}

    \item \textbf{ShapLength \cite{mariotti2022measuring}}: ShapLength is a model-agnostic and computationally efficient metric for assessing how human-understandable a model is. It builds upon the p\%-complete explanation property, which finds the smallest set of features whose SHAP values sum exceed a defined threshold. Shap Length then simply represents the number of features included in this p\%-complete explanation. A higher ShapLength indicate a model that relies on a larger number of features or complex interactions, potentially making it harder to explain and interpret.
\end{enumerate}

\section{Experimental Results}\label{Results}
In this section, we quantitatively evaluate the impact of DP on the effectiveness (Subsection \ref{DPvsAD}) and explainability of the AD models. Explainability is assessed through quantitative and qualitative evaluations, in subsection \ref{DPvsXAIQuant} and \ref{DPvsXAIQual}, respectively.
\subsection{Impact of Differential Privacy on Anomaly Detection Models}\label{DPvsAD}
\begin{table}[h]
\centering
\caption{AUC and precision of iForest across the mammography, thyroid, and bank datasets, without DP and with varying DP-$\varepsilon$ values, considering the two noise-adding mechanisms: Gaussian and Laplace.}
\label{AUCPREC:iForest}
\begin{tabular}{c|c|c|cccc|cccc}
\multirow{3}{*}{Dataset} & \multirow{3}{*}{Metric} & \multirow{3}{*}{Without Privacy} & \multicolumn{4}{c|}{Laplace} & \multicolumn{4}{c}{Gaussian} \\ \cline{4-11} 
 &  &  & \multicolumn{4}{c|}{$\varepsilon$} & \multicolumn{4}{c}{$\varepsilon$} \\
 &  &  & \multicolumn{1}{c|}{5} & \multicolumn{1}{c|}{1} & \multicolumn{1}{c|}{0.1} & 0.01 & \multicolumn{1}{c|}{5} & \multicolumn{1}{c|}{1} & \multicolumn{1}{c|}{0.1} & 0.01 \\ \hline
\multirow{2}{*}{mammography} & AUC & 74 & \multicolumn{1}{c|}{73} & \multicolumn{1}{c|}{72} & \multicolumn{1}{c|}{66} & 53 & \multicolumn{1}{c|}{76} & \multicolumn{1}{c|}{72} & \multicolumn{1}{c|}{69} & 54 \\
 & Precision & 90 & \multicolumn{1}{c|}{91} & \multicolumn{1}{c|}{90} & \multicolumn{1}{c|}{88} & 84 & \multicolumn{1}{c|}{92} & \multicolumn{1}{c|}{90} & \multicolumn{1}{c|}{89} & 85 \\ \hline
\multirow{2}{*}{thyroid} & AUC & 89 & \multicolumn{1}{c|}{54} & \multicolumn{1}{c|}{56} & \multicolumn{1}{c|}{51} & 50 & \multicolumn{1}{c|}{54} & \multicolumn{1}{c|}{53} & \multicolumn{1}{c|}{53} & 48 \\
 & Precision & 90 & \multicolumn{1}{c|}{58} & \multicolumn{1}{c|}{60} & \multicolumn{1}{c|}{56} & 55 & \multicolumn{1}{c|}{58} & \multicolumn{1}{c|}{58} & \multicolumn{1}{c|}{58} & 53 \\ \hline
\multirow{2}{*}{bank} & AUC & 64 & \multicolumn{1}{c|}{58} & \multicolumn{1}{c|}{57} & \multicolumn{1}{c|}{52} & 52 & \multicolumn{1}{c|}{57} & \multicolumn{1}{c|}{56} & \multicolumn{1}{c|}{51} & 49 \\
 & Precision & 68 & \multicolumn{1}{c|}{64} & \multicolumn{1}{c|}{63} & \multicolumn{1}{c|}{58} & 58 & \multicolumn{1}{c|}{63} & \multicolumn{1}{c|}{62} & \multicolumn{1}{c|}{58} & 55
\end{tabular}
\end{table}
\begin{table}[h]
\centering
\caption{AUC and precision of LOF across the mammography, thyroid, and bank datasets, without DP and with varying DP-$\varepsilon$ values, considering the two noise adding mechanism: Gaussian and Laplace.}
\label{AUCPREC:LOF}
\begin{tabular}{c|c|c|cccc|cccc}
\multirow{3}{*}{Dataset} & \multirow{3}{*}{Metric} & \multirow{3}{*}{Without Privacy} & \multicolumn{4}{c|}{Laplace} & \multicolumn{4}{c}{Gaussian} \\ \cline{4-11} 
 &  &  & \multicolumn{4}{c|}{Epsilon} & \multicolumn{4}{c}{Epsilon} \\
 &  &  & \multicolumn{1}{c|}{5} & \multicolumn{1}{c|}{1} & \multicolumn{1}{c|}{0.1} & 0.01 & \multicolumn{1}{c|}{5} & \multicolumn{1}{c|}{1} & \multicolumn{1}{c|}{0.1} & 0.01 \\ \hline
\multirow{2}{*}{Mammography} & AUC & 74 & \multicolumn{1}{c|}{74} & \multicolumn{1}{c|}{74} & \multicolumn{1}{c|}{74} & 66 & \multicolumn{1}{c|}{74} & \multicolumn{1}{c|}{74} & \multicolumn{1}{c|}{73} & 70 \\
 & Precision & 91 & \multicolumn{1}{c|}{91} & \multicolumn{1}{c|}{91} & \multicolumn{1}{c|}{91} & 88 & \multicolumn{1}{c|}{91} & \multicolumn{1}{c|}{91} & \multicolumn{1}{c|}{91} & 89 \\ \hline
\multirow{2}{*}{Thyroid} & AUC & 56 & \multicolumn{1}{c|}{56} & \multicolumn{1}{c|}{56} & \multicolumn{1}{c|}{53} & 51 & \multicolumn{1}{c|}{56} & \multicolumn{1}{c|}{57} & \multicolumn{1}{c|}{53} & 51 \\
 & Precision & 60 & \multicolumn{1}{c|}{60} & \multicolumn{1}{c|}{60} & \multicolumn{1}{c|}{58} & 56 & \multicolumn{1}{c|}{60} & \multicolumn{1}{c|}{61} & \multicolumn{1}{c|}{58} & 56 \\ \hline
\multirow{2}{*}{Bank} & AUC & 59 & \multicolumn{1}{c|}{59} & \multicolumn{1}{c|}{59} & \multicolumn{1}{c|}{59} & 59 & \multicolumn{1}{c|}{59} & \multicolumn{1}{c|}{59} & \multicolumn{1}{c|}{59} & 59 \\
 & Precision & 65 & \multicolumn{1}{c|}{65} & \multicolumn{1}{c|}{65} & \multicolumn{1}{c|}{64} & 65 & \multicolumn{1}{c|}{65} & \multicolumn{1}{c|}{64} & \multicolumn{1}{c|}{64} & 64
\end{tabular}%
\end{table}

We start by analyzing the impact of employing DP on the performance of iForest and LOF. Table \ref{AUCPREC:iForest} reports the AUC and precision of iForest across the three considered datasets (mammography, thyroid, and bank), and across the two noise-adding mechanisms (Gaussian and Laplace) for varying values of $\varepsilon$.
When DP is not employed, iForest achieves an AUC of 74\%, 89\%, and 64\% for mammography, thyroid, and bank datasets, respectively. A precision of 90\% for both the mammography and thyroid datasets, and 68\% for the bank dataset.
As DP is introduced, iForest generally exhibits decreased performance compared to the non-DP models. AUC and precision decrease already for high values of $\varepsilon$ (i.e., less privacy). As expected, the difference is higher for smaller values of $\varepsilon$ (i.e., more privacy). By decreasing $\varepsilon$ in the Laplace case, the AUC decreases from 73\% to 53\% for the mammography dataset, from 54\% to 50\% for the thyroid dataset, and from 58\% to 52\% for the bank dataset, and a similar decrease happens when decreasing $\varepsilon$ with Gaussian noise. This is except for one case, that is for the mammography dataset for $\varepsilon$ = 5 with Gaussian.

Table \ref{AUCPREC:LOF} reports the AUC and precision achieved by LOF across three considered datasets. Results show that in the case without applying DP, LOF achieves an AUC of 74\%, 56\%, and 59\% for mammography, thyroid, and bank datasets, respectively, and a precision of 91\%, 60\%, and 65\%. As DP is introduced, unlike with iForest, we observe a similar value for both the AUC and the precision across all datasets and all values of $\varepsilon$ except for $\varepsilon$ = 0.01. 

This suggests that while iForest initially outperformed LOF without DP, LOF proved to be more robust and resilient to DP, maintaining its effectiveness under such constraints better than iForest, potentially due to its focus on k-nearest neighbors and local data density. This investigation suggests a trade-off between local and global outlier detection capabilities under DP. We explore this trade-off further with respect to SHAP explanations.

\subsection{Impact of Differential Privacy on SHAP explanations}\label{DPvsXAIQuant}

\begin{figure}[]
    \centering
    \begin{subfigure}{.57\linewidth}
        \includegraphics[width=\linewidth]{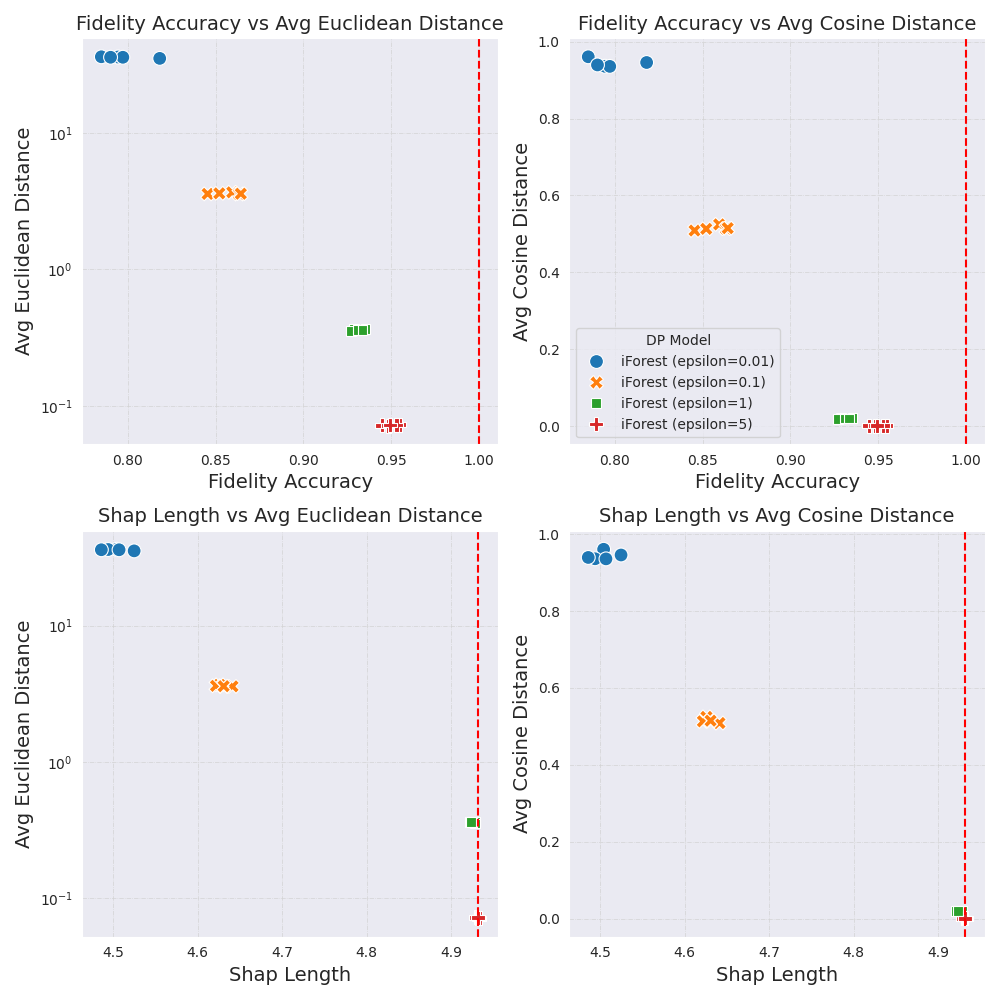}
        \caption{Mammography dataset}
        \label{Mammo Laplace SHapGAP}
    \end{subfigure}%
    \begin{subfigure}{.57\linewidth}
        \includegraphics[width=\linewidth]{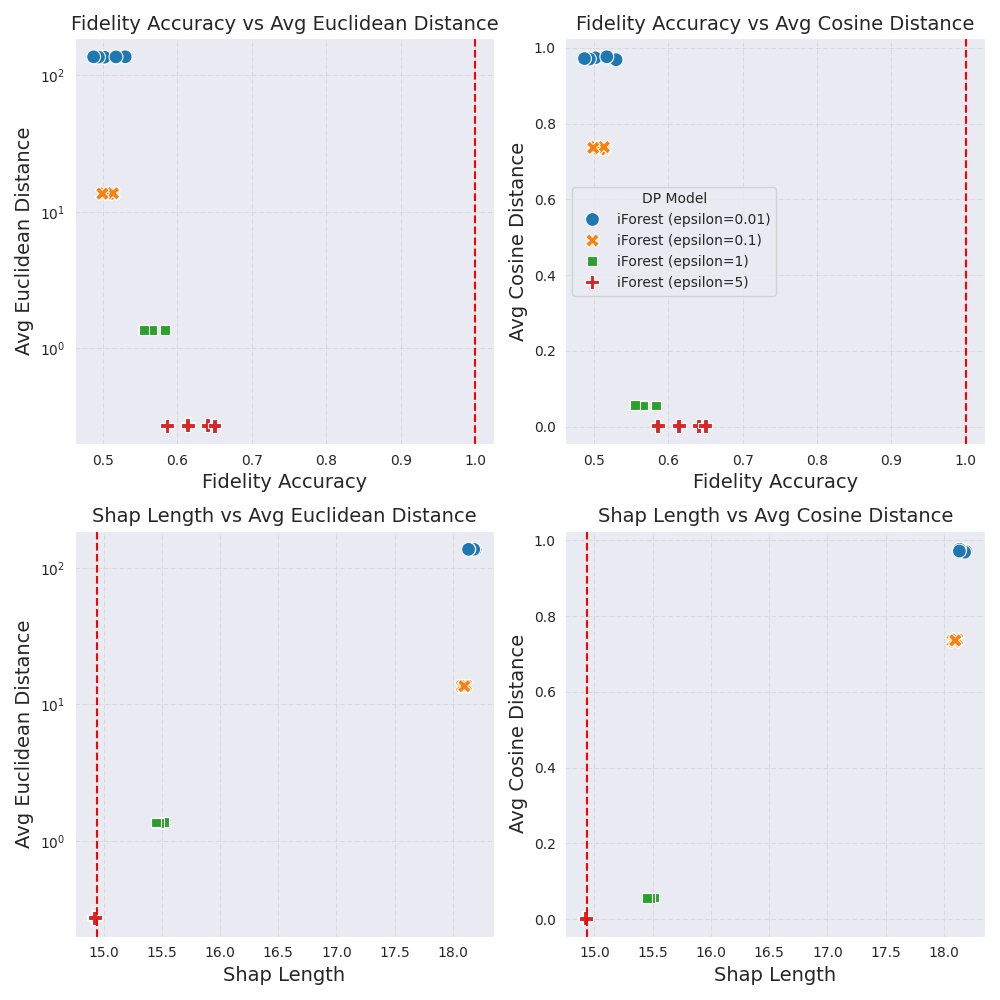}
        \caption{Thyroid dataset}
        \label{Thyroid Laplace SHapGAP}
    \end{subfigure}
    
    \begin{subfigure}{.57\linewidth}
        \includegraphics[width=\linewidth]{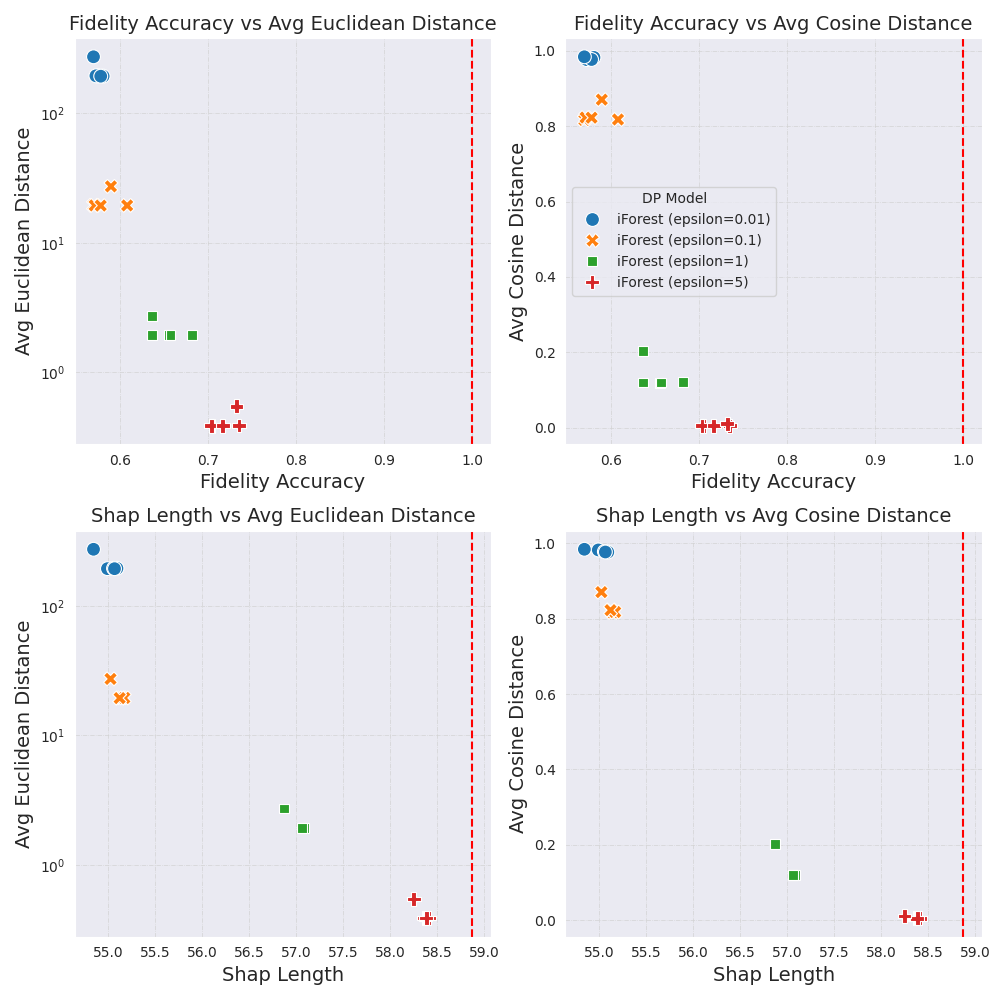}
        \caption{Bank dataset}
        \label{bank Laplace SHapGAP}
    \end{subfigure}
    \caption{Fidelity Accuracy of iForest and average ShapGap-Euclidean distance, ShapGap-Cosine distance and ShapLength computed across the explanations extracted using SHAP for the various iForest models and the various values of epsilon, across (a) Mammography, (b) Thyroid and (c) Bank datasets. The vertical dashed line represents the without DP metric presented at the x-axis.}
    \label{fig:shapgapiForest}
\end{figure}

Figures \ref{fig:shapgapiForest} and \ref{fig:shapgapLOF} report the results of ShapGap-Cosine and ShapGap-Euclidean distances along with the fidelity accuracy and ShapLength metrics of iForest and LOF, across the three datasets and for the various values of $\varepsilon$ considered. Each row of the plot focuses on a specific metric on the x-axis. The y-axis consistently displays both ShapGap-Cosine and ShapGap-Euclidean distances across 5 runs for each $\varepsilon$ experiment. Each point on the plots corresponds to the average of the corresponding ShapGap distance across all data points of each dataset for one single run. In an ideal scenario, the AD models with and without DP should agree to 100\%, producing identical outputs for all data points. 

\subsubsection{iForest:}\label{iForestGapAnalysis}
Figure \ref{fig:shapgapiForest}, presents the results for iForest across the three datasets. Results show that ShapGAP-Euclidean and ShapGap-Cosine distances across all datasets increase as the privacy guarantee increases (i.e., $\varepsilon$ decreases). This difference means that the vectors of the features in the explanations extracted using SHAP before and after the application of DP change in both magnitude (captured by the euclidean distance) and direction (captured by the cosine distance). These findings reveal that as the privacy parameter $\varepsilon$ decreases, the magnitude of SHAP values tends to deviate more from those obtained in the absence of privacy constraints, for instance, a ShapGap-Cosine value close to 1 indicates high dissimilarity in the magnitude and direction of the SHAP value vectors before and after applying DP with $\varepsilon$ of 0.01. Conversely, ShapGap-Euclidean has no upper bound, with higher values signifying greater dissimilarity. Specifically, the ShapGap-Euclidean metric for the mammography dataset ranges between 0 and 10, while for the thyroid and bank dataset, it ranges between 0 and 100. 
In contrast, the ShapGap-Cosine metric (Figure \ref{fig:shapgapiForest} a,b,c bottom and up right) scored between 0 to 1 for both datasets depending on $\varepsilon$. The results also show a consistent trend between the value of $\varepsilon$ and the fidelity accuracy (Figure. \ref{fig:shapgapiForest} a,b,c up right), as ShapGap-Cosine with smaller $\varepsilon$ values correspond to lower fidelity accuracy, progressively decreasing from 100\% relative to the ideal scenario when DP is not applied. This trend is evident for both distances (Figure. \ref{fig:shapgapiForest} a,b,c up) and across all the datasets.

Another notable trend is the negative correlation between fidelity scores and distance values (i.e., higher distances imply lower fidelity scores, and vice versa). In other words, when the model's reasoning aligns more closely with the original non-DP model (higher fidelity scores), the distances between SHAP values are minimized, indicating a stronger agreement in the influence of features on predictions. This trend is observable for both ShapGap-Euclidean and cosine distances, showing that both the magnitude and direction of the SHAP value vectors are closely linked to the model's reasoning and fidelity. \textit{This result suggests that there is a correlation between the difference in explanations (ShapGap) and the model's ability to faithfully represent the original model's predictions, as indicated by the fidelity scores}.

We now discuss the complexity captured by the value of ShapLength. The findings vary based on the privacy budget ($\varepsilon$) value for DP and depending on the dataset. For the mammography dataset (Figure \ref{fig:shapgapiForest}a bottom), observations at $\varepsilon$ values of 1 and 5 indicate stability and consistency in ShapLength, maintaining an average value of 4.95, identical to that of the model without DP. \textit{This implies that the implementation of DP, while ensuring a moderate level of privacy protection, does not impact the ShapLength, preserving the explainability complexity of the model as in the non-DP setting}. However, at lower $\varepsilon$ values, specifically 0.1 and 1, there is a noticeable reduction in ShapLength to 4.5, indicating a slight deviation in model complexity compared to the model without DP. This indicates that DP with stricter privacy has reduced the model's complexity. For the thyroid dataset (Figure \ref{fig:shapgapiForest}b bottom), we observe a different outcome: the ShapLength is highly affected by the application of DP, as with the decrease of $\varepsilon$, the ShapLength is increasing. When observing it with the SHAP Gap distances, smaller ShapLength aligns with larger Euclidean and Cosine distances. The bank dataset (Figure \ref{fig:shapgapiForest}c bottom) exhibits a similar outcome in ShapLength compared to the mammography, as we observe a decrease in SHAP Length with smaller $\varepsilon$ values even with $\varepsilon$ = 5 where we see that the ShapLength has also decreased.

Our investigation reveals a privacy-explainability trade-off in applying DP to iForest models with SHAP explanations. Stricter privacy (lower $\varepsilon$) leads to increased divergence in SHAP values (magnitude and direction), decreased fidelity to the original model, and potentially simpler models with less detailed explanations (reduced ShapLength) and are dependent. Conversely, relaxed privacy settings show better-preserved interpretability with explanations closer to the non-DP model and stable model complexity. \textit{This negative correlation between explanation divergence and fidelity scores suggests a link between interpretability and the model's ability to faithfully represent the original model's predictions and explanations.}

\subsubsection{LOF:}
\begin{figure}[]
\centering
    \begin{subfigure}{.57\linewidth}
        \includegraphics[width=\linewidth]{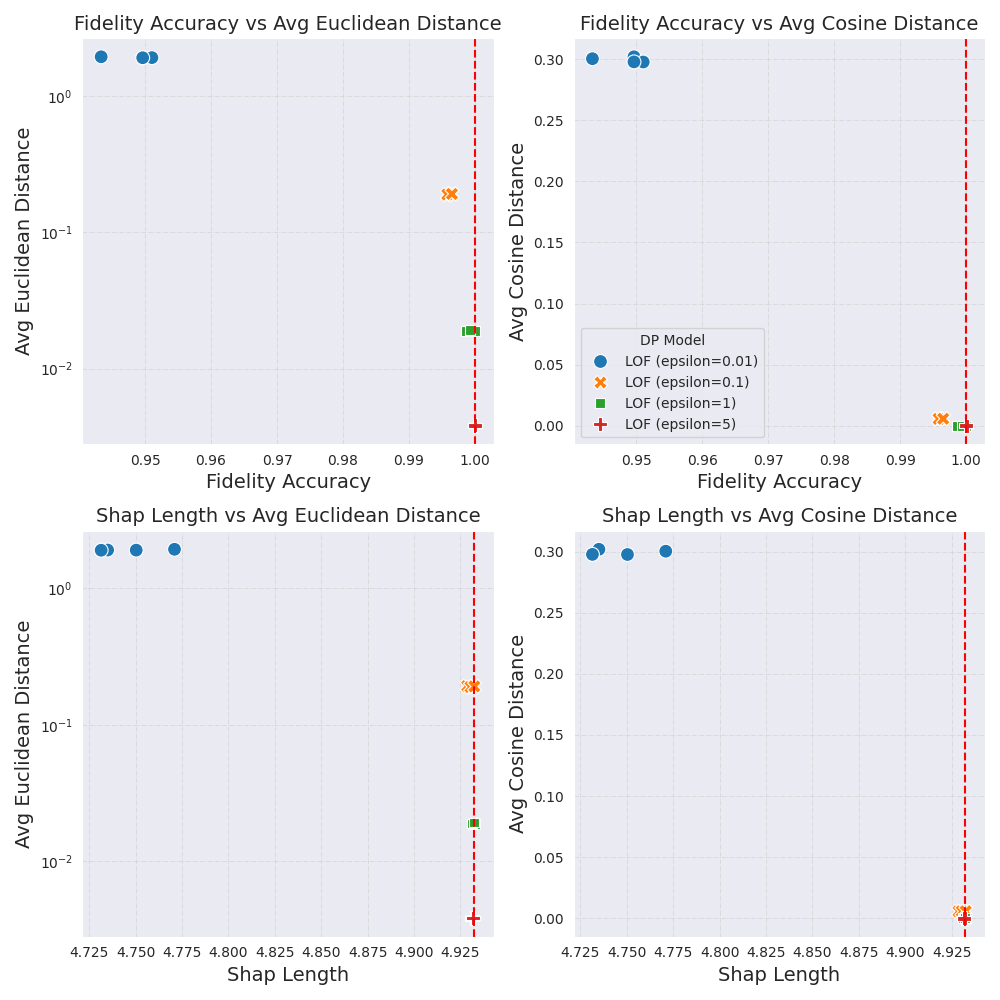}
        \caption{Mammography dataset}
        \label{Mammo Laplace SHapGAP}
    \end{subfigure}%
    \begin{subfigure}{.57\linewidth}
        \includegraphics[width=\linewidth]{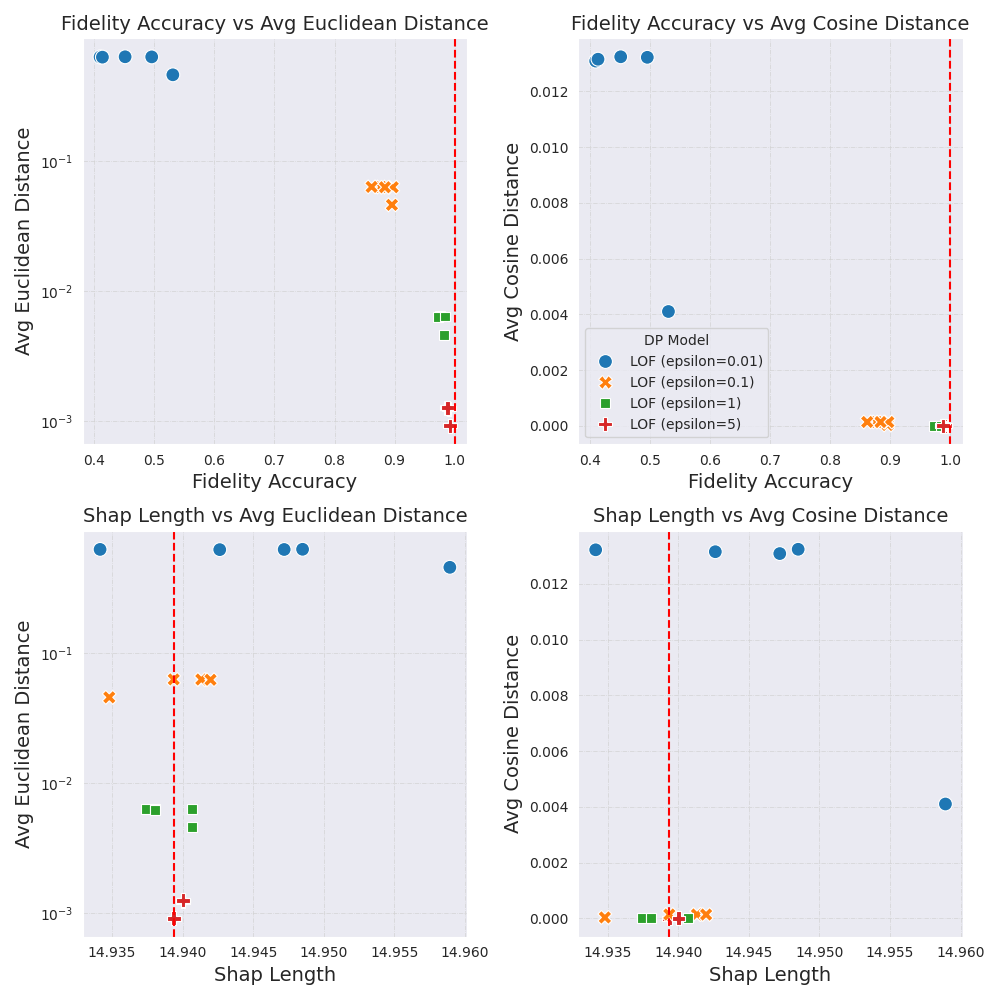}
        \caption{Thyroid dataset}
        \label{Thyroid Laplace SHapGAP}
    \end{subfigure}
    
    \begin{subfigure}{.55\linewidth}
        \includegraphics[width=\linewidth]{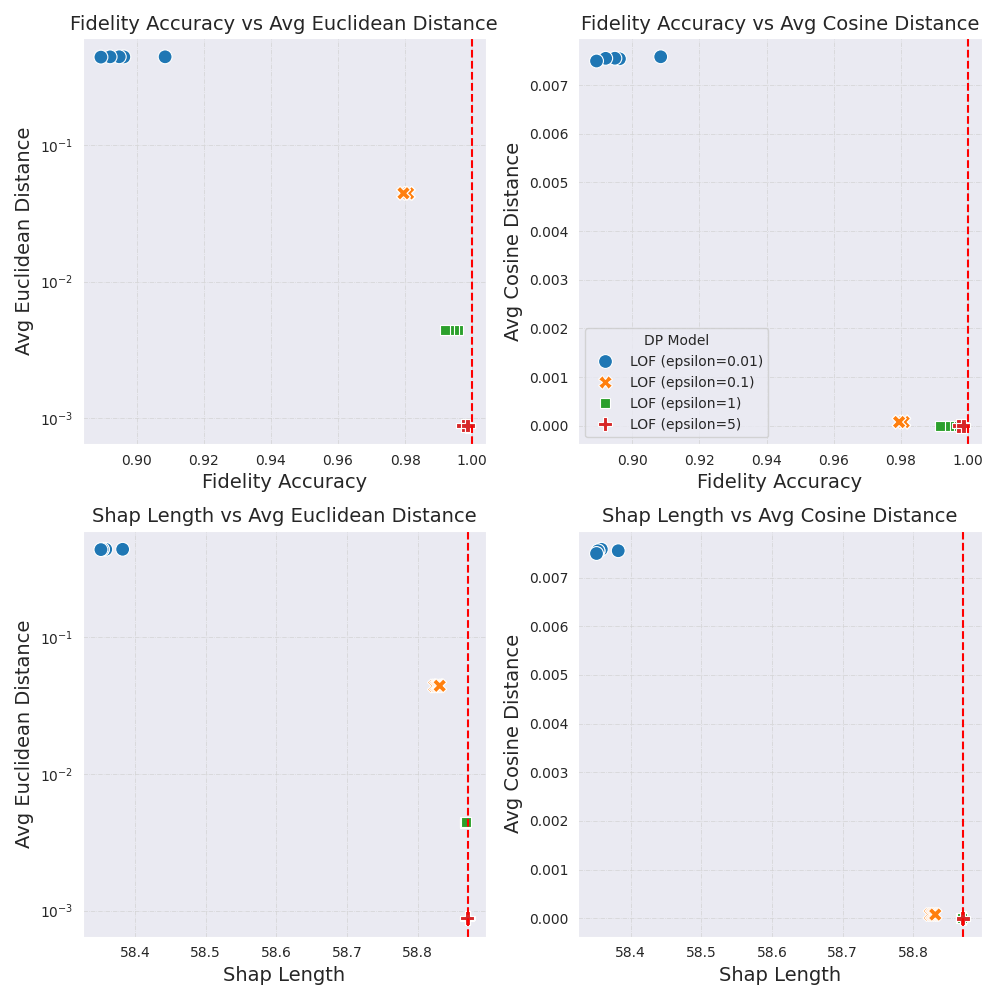}
        \caption{Bank dataset}
        \label{bank Laplace SHapGAP}
    \end{subfigure}
    \caption{Fidelity Accuracy of LOF and average ShapGap-Euclidean distance, ShapGap-Cosine distance, and ShapLength computed across the explanations extracted using SHAP for the various iForest models and the various values of $\varepsilon$, across (a) Mammography, (b) Thyroid and (c) Bank datasets. The vertical dashed line represents the without DP metric presented at the x-axis.}
    \label{fig:shapgapLOF}
\end{figure}
Figure \ref{fig:shapgapLOF} shows the ShapGap metrics for the performance of the LOF model, highlighting the association between model fidelity and privacy levels achieved through DP (Figure \ref{fig:shapgapLOF} a,b,c up). Across the mammography, thyroid, and bank datasets, model fidelity remains impressively high, ranging from 90\% to 100\% for $\varepsilon$ values of 0.1, 1, and 5, nearly mirroring the fidelity seen in the AD model without DP. However, a notable fidelity reduction occurs at 
$\varepsilon=0.01$, with drops by approximately 5\%, 40\%, and 10\% for the mammography, thyroid, and bank datasets, respectively. Despite this high fidelity, we observe shifts and increases in Euclidean distances (Figure \ref{fig:shapgapLOF} a,b,c up left), indicating alterations in the scale of the underlying data that fidelity scores do not capture. Conversely, the ShapGap-Cosine distances (Figure \ref{fig:shapgapLOF} a,b,c up right), remain largely stable across 0.1, 1, and 5, $\varepsilon$ values, suggesting that the directionality of the SHAP value vectors stays consistent despite the application of DP except for $\varepsilon$ of 0.01. Specifically, for the mammography dataset, it increases to around 0.3. For the thyroid, and bank dataset, the ShapGap-Cosine also increases but stays within a very low cosine measure of around 0.07 and 0.012 respectively, indicating a low magnitude and direction of change within SHAP values. \textit{This phenomenon suggests that while the overall magnitude of model explanations is influenced by DP, the vector direction reflecting feature contributions towards predictions remains minimally unaffected with DP for LOF.}

Analyzing model complexity through the ShapLength (Figure \ref{fig:shapgapLOF} a,b,c bottom), we find that in both the mammography and bank datasets, ShapLength remains relatively unchanged for $\varepsilon$ values of 0.1, 1, and 5, suggesting minimal variations in model complexity. However, at a $\varepsilon$ of 0.01, we notice a decline in ShapLength, which decreases from 4.925 to 4.725 in the mammography dataset and from 58.8 to 58.4 in the bank dataset. This indicates that the complexity of the model decreases slightly under more strict privacy conditions. In contrast, the thyroid dataset demonstrates a different pattern: ShapLength stays closely aligned with the small variance in complexity without DP, except at a $\varepsilon$ of 0.01, where we note a relatively small increase in the variance of complexity as the privacy level increases across the different runs (from 14.935 to 14.96). 

The observed inconsistencies across datasets in fidelity and ShapGap-Euclidean and Cosine distances likely stem from the inherent randomness introduced by DP and the chosen AD algorithms. While DP guarantees privacy, its added noise can alter various data characteristics, therefore the noise level and privacy level should be carefully chosen in a way that the overall distribution of the data is not largely affected. Therefore, this means that fidelity scores can remain elevated despite variations in Euclidean distance within higher $\varepsilon$. This also suggests that the data distribution undergoes modifications caused by the DP noise, but these modifications are limited such that the overall statistical properties remain largely preserved, thus maintaining high fidelity and low ShapGap-Cosine.

As anticipated, these findings indicate that iForest is more sensitive to data points that significantly deviate from the overall distribution of the data, while LOF more easily detects deviations within specific data regions. Since DP focuses on preserving overall data statistics, it can alter the distribution of the data, impacting how anomalies appear in the global picture. This, in turn, affects the SHAP values in iForest, as features contributing to isolation might be masked by the DP noise. LOF focuses on Local outliers and analyzes how different a data point is from its closest neighbors (local perspective). DP's impact on local neighborhoods might be less significant compared to its effect on the entire data distribution. Additionally, LOF's SHAP values might focus on features relevant to the local anomaly score, which might be less sensitive to global distribution changes caused by DP.
\subsection{ShapGap Distribution Analysis}\label{BoxPlots}
To further explore the distribution of SHAP explanation divergence across data points, we report the findings using box plots for both ShapGap-Euclidean and ShapGap-Cosine distances.

\begin{figure}[h]
    \centering
    \begin{subfigure}{.35\linewidth}
        \includegraphics[width=\linewidth,height=120pt]{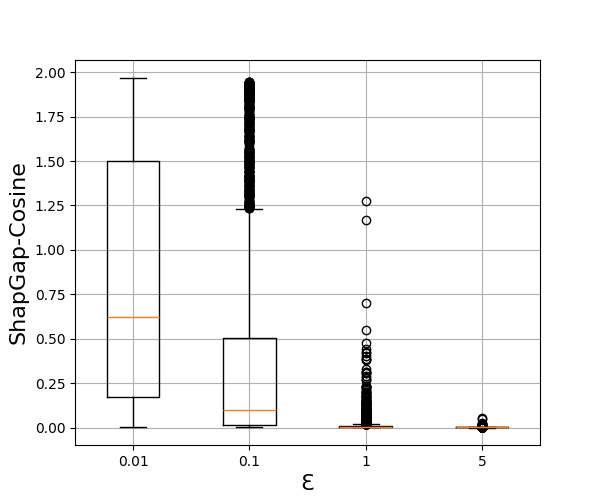}
        \caption{Mammography dataset}
    \end{subfigure}%
    \begin{subfigure}{.35\linewidth}
        \includegraphics[width=\linewidth,height=120pt]{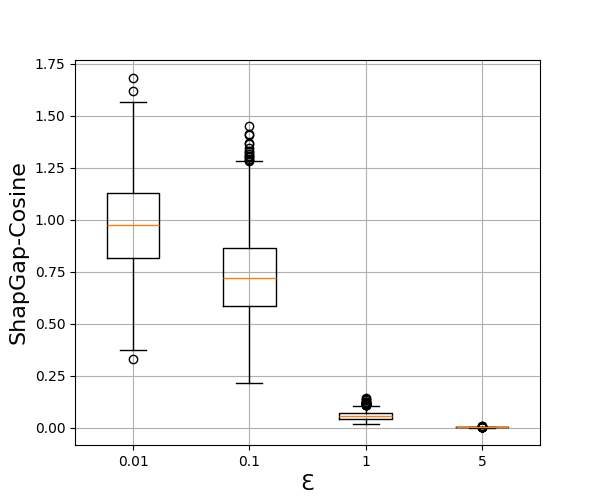}
        \caption{Thyroid dataset}
    \end{subfigure}%
    \begin{subfigure}{.35\linewidth}
        \includegraphics[width=\linewidth,height=120pt]{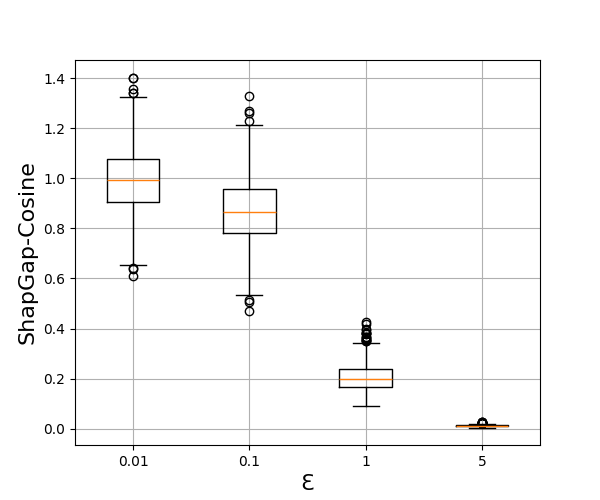}
        \caption{Bank dataset}
    \end{subfigure}
    \caption{Distribution of iForest ShapGap-Cosine distances across a) Mammography, b) Thyroid, and c) Bank Datasets for the various $\varepsilon$ values}
    \label{fig:boxiForestCosine}
\end{figure}

\begin{figure}[h]
    \centering
    \begin{subfigure}{.35\linewidth}
        \includegraphics[width=\linewidth,height=120pt]{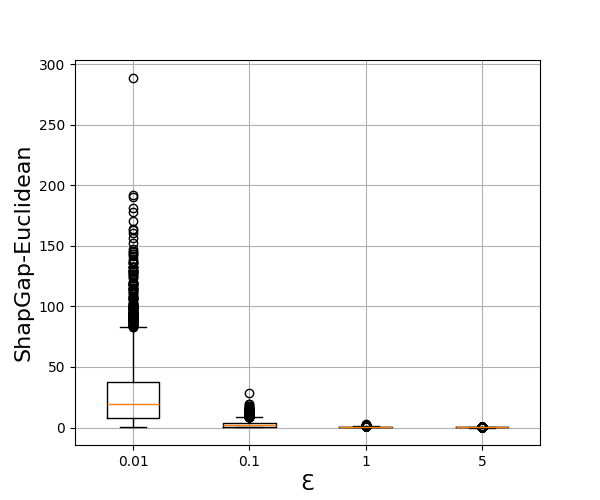}
        \caption{Mammography dataset}
    \end{subfigure}%
    \begin{subfigure}{.35\linewidth}
        \includegraphics[width=\linewidth,height=120pt]{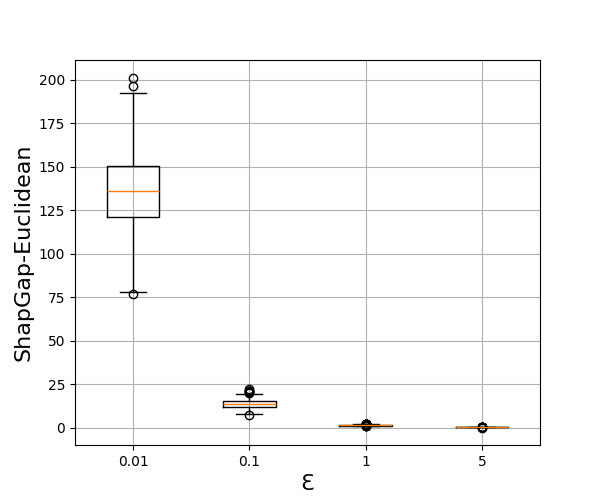}
        \caption{Thyroid dataset}
    \end{subfigure}%
    \begin{subfigure}{.35\linewidth}
        \includegraphics[width=\linewidth, height=120pt]{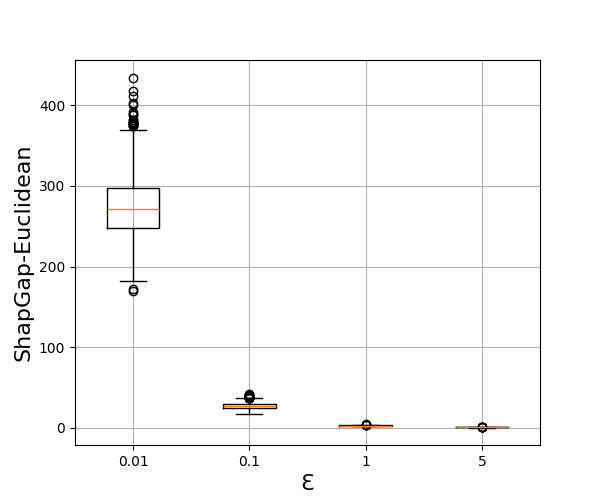}
        \caption{Bank dataset}
    \end{subfigure}%
    \caption{Distribution of iForest ShapGap-Euclidean distances across a) Mammography, b) Thyroid, and c) Bank Datasets for the various $\varepsilon$ values}
    \label{fig:boxiForestEuclidean}
\end{figure}

In Figure \ref{fig:boxiForestCosine}, we visualize the distribution of ShapGap-Cosine for iForest for all data points across the three datasets. We observe that for $\varepsilon$ values of 1 and 5 across the three datasets, the ShapGap-Cosine has a lower spread in comparison to smaller $\varepsilon$ of 0.1 and 0.01. This low spread indicates that most of the data points have a small cosine distance between 0 and 0.25, which means closer to the non-DP scenario. However, for smaller $\varepsilon$ values (0.1 and 0.01), we see a wider spread of the ShapGap-Cosine distribution. The values range between 0 and 2 for $\varepsilon$ = 0.01 with the mammography dataset, between 0.25 and 1.75 for thyroid, and between 0.6 and 1.4 for bank. This wider distribution suggests a significant portion of data points exhibiting high ShapGap, deviating from the behavior observed without DP. Regarding Euclidean distances, Fig. \ref{fig:boxiForestEuclidean} we visualize the distribution of ShapGap-Euclidean for iForest across all data points for each dataset. We observe that for $\varepsilon$ values of 0.1, 1, and 5 across the three datasets, the ShapGap Euclidean distance has a small distribution compared to $\varepsilon$ of 0.01. This indicates that the data points with small Euclidean distances (ranging between 0 and 20 for mammography, from 0 to 25 for thyroid, and from 0 to 50 for the bank), are closer to the non-DP scenario. However, for smaller $\varepsilon$ values (0.01), we note a larger distribution of ShapGap Euclidean measures reaching up to 500 in the bank dataset, which is extremely large. 
\vspace{0mm}
\vspace{0mm}
\begin{figure}[h]
    \centering
    \begin{subfigure}{.35\linewidth}
        \includegraphics[width=\linewidth, height=110pt]{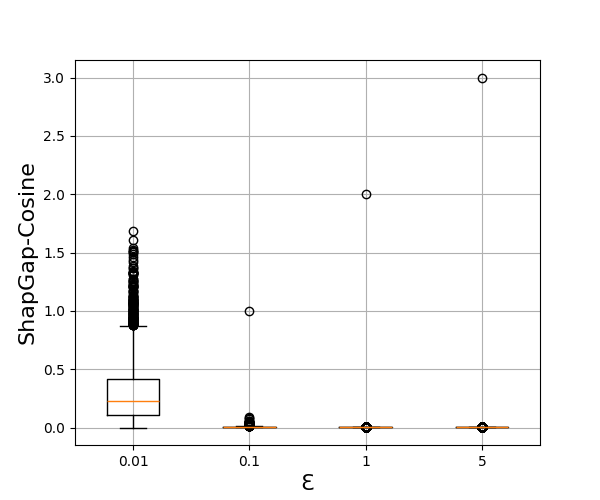}
        \caption{Mammography dataset}
    \end{subfigure}%
    \begin{subfigure}{.35\linewidth}
        \includegraphics[width=\linewidth, height=110pt]{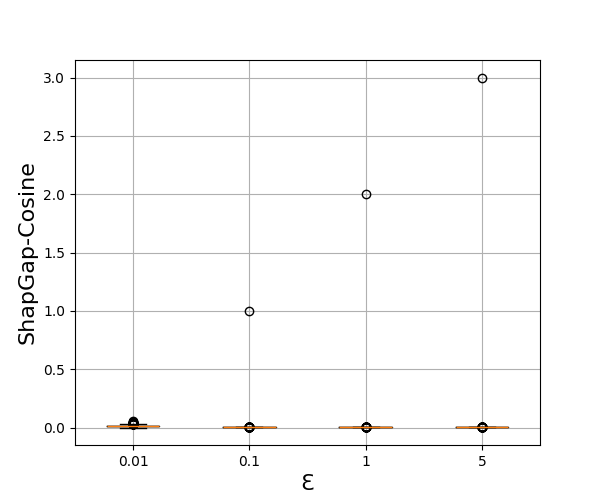}
        \caption{Thyroid dataset}
    \end{subfigure}%
    \begin{subfigure}{.35\linewidth}
        \includegraphics[width=\linewidth, height=110pt]{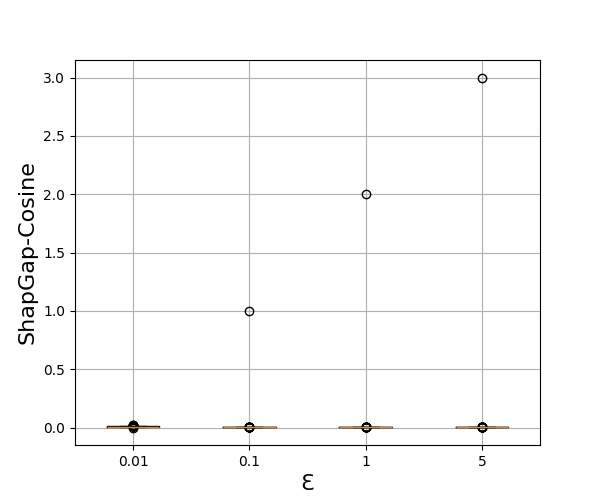}
        \caption{Bank dataset}
    \end{subfigure}
    \caption{Distribution of LOF ShapGap-Cosine distances across a) Mammography, b) Thyroid, and c) Bank Datasets for the various $\varepsilon$ values}
    \label{fig:boxLOFCosine}
\end{figure}\vspace{0mm}
\begin{figure}[h]
    \centering
    \begin{subfigure}{.35\linewidth}
        \includegraphics[width=\linewidth, height=110pt]{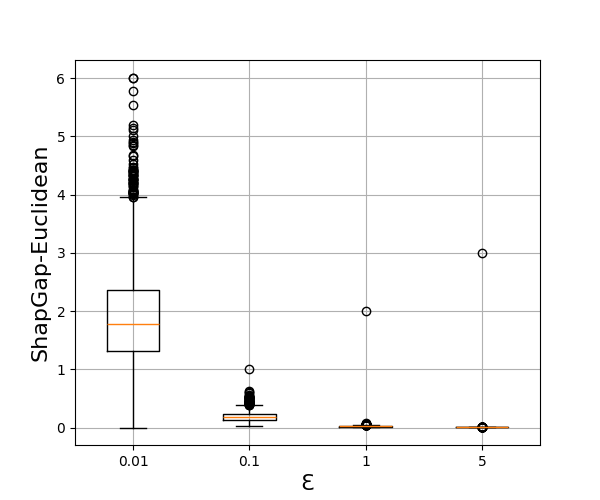}
        \caption{Mammography dataset}
    \end{subfigure}%
    \begin{subfigure}{.35\linewidth}
        \includegraphics[width=\linewidth, height=110pt]{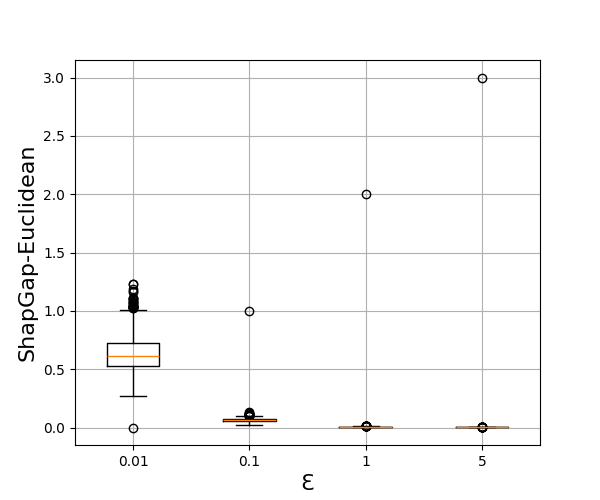}
        \caption{Thyroid dataset}
    \end{subfigure}%
    \begin{subfigure}{.35\linewidth}
        \includegraphics[width=\linewidth, height=110pt]{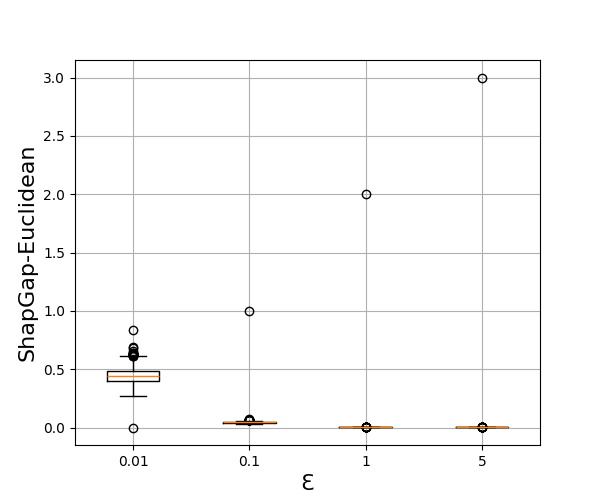}
        \caption{Bank dataset}
    \end{subfigure}%
    \caption{Distribution of LOF ShapGap-Euclidean distances across a) Mammography, b) Thyroid, and c) Bank Datasets for the various $\varepsilon$ values}
    \label{fig:boxLOFEuclidean}
\end{figure}
Figure \ref{fig:boxLOFCosine} shows the distribution of ShapGap Cosine across all data points with LOF. For $\varepsilon$ values of 0.1, 1, and 5, the distribution of ShapGap Cosine distance is relatively small (less than 0.2 for all the datasets), indicating that the SHAP values with and without DP are similar for most of the data points, suggesting minimal impact on the data due to DP at these privacy levels. For $\varepsilon$ equal to 0.01, the distribution of ShapGap Cosine distance is larger (ranging from 0 to 1.75), indicating greater differences between SHAP values with and without DP. Regarding Euclidean distances, Figure \ref{fig:boxLOFEuclidean} visualizes the distribution of ShapGap Euclidean for LOF across all data points. We observe that for $\varepsilon$ of 0.1, 1, and 5, for the 3 datasets, the ShapGap Euclidean concentrates around at most between 0 and 1, and there is no diverged distribution. While only for the very small $\varepsilon$ we observe the distribution of ShapGap Euclidean diverges covering a range between 0 and 6 as maximum, which is still relatively very small compared to the magnitude change scale of iForest. 

These findings suggest that with moderate privacy levels have minimal impact on SHAP with LOF. However, iForest appears to be more sensitive to the noise introduced by DP, evidenced by the large distribution of ShapGap distances and reflected by the distance distribution across the data points.

\subsection{Impact of Differential Privacy on SHAP summary plots}\label{DPvsXAIQual}
\begin{figure}[h]
    \centering
    \begin{subfigure}{.55\linewidth}
        \includegraphics[width=\linewidth,height=90pt]{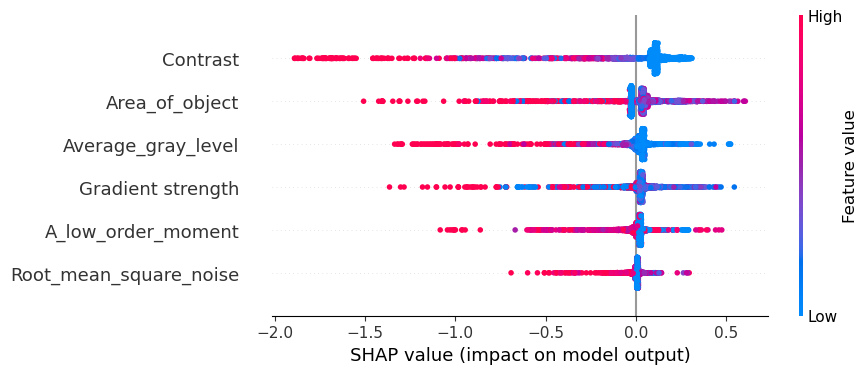}
        \caption{without DP}
        \label{mammo iso}
    \end{subfigure}%
    \begin{subfigure}{.55\linewidth}
        \includegraphics[width=\linewidth,height=90pt]{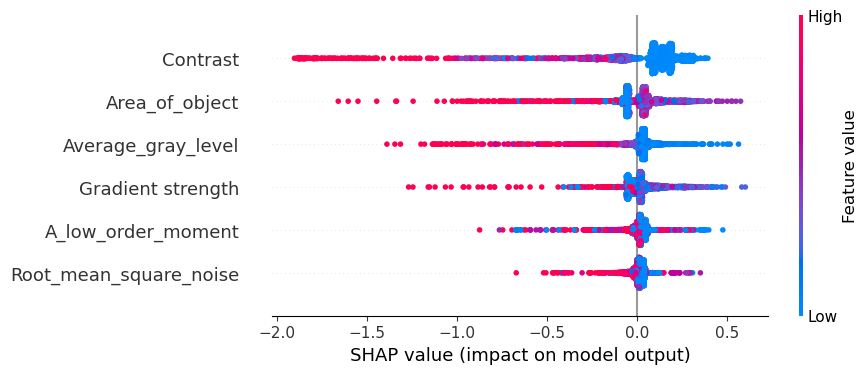}
        \caption{$\varepsilon$=5}
        \label{laplace mammo iso e=5}
    \end{subfigure}%
    
    \begin{subfigure}{.55\linewidth}
        \includegraphics[width=\linewidth,height=90pt]{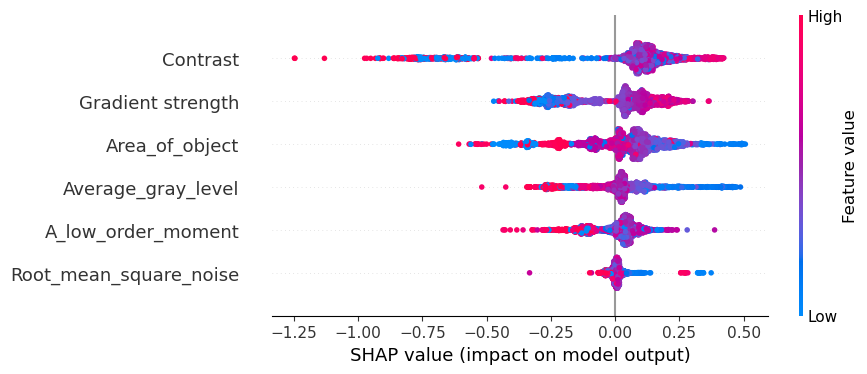}
        \caption{$\varepsilon$=0.01}
        \label{laplace mammo iso e=0.01}
    \end{subfigure}
    \caption{Summary plot for iForest for the mammography dataset with various $\varepsilon$}
    \label{fig:MammoIforestLaplace}
\end{figure}

\begin{figure}[h]
    \centering
    \begin{subfigure}{.55\linewidth}
        \includegraphics[width=\linewidth, height=90pt]{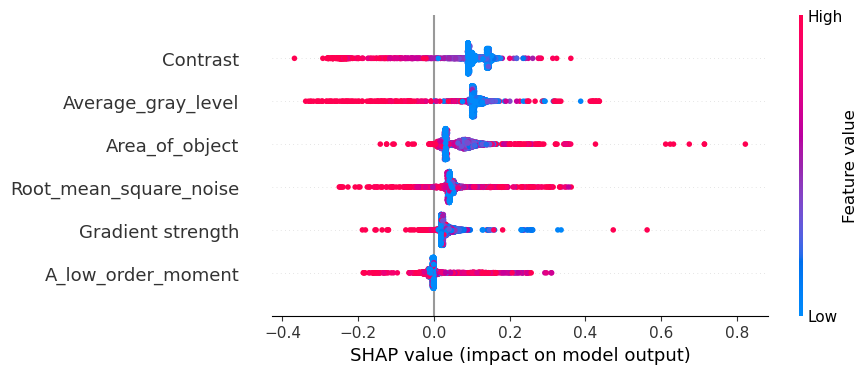}
        \caption{without DP}
        \label{mammo iso}
    \end{subfigure}%
    \begin{subfigure}{.55\linewidth}
        \includegraphics[width=\linewidth, height=90pt]{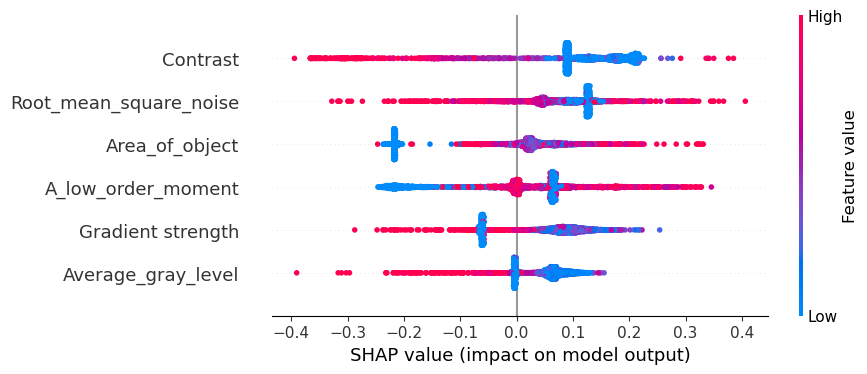}
        \caption{$\varepsilon$=5}
        \label{laplace mammo lof e=5}
    \end{subfigure}
    
    \begin{subfigure}{.55\linewidth}
        \includegraphics[width=\linewidth, height=90pt]{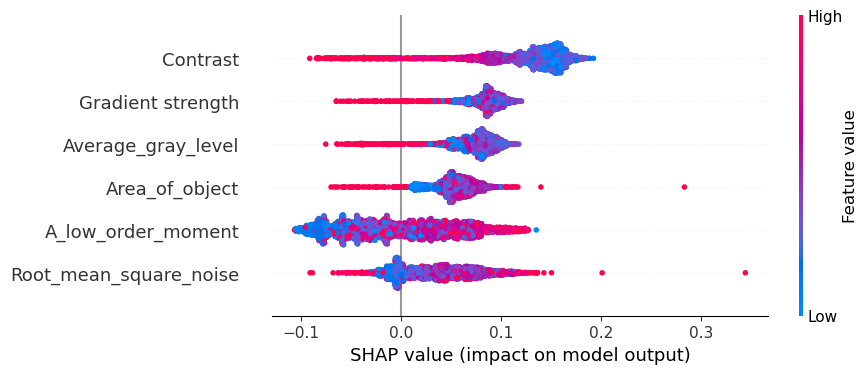}
        \caption{$\varepsilon$=0.01}
        \label{laplace mammo lof e=0.01}
    \end{subfigure}
    \caption{Summary plot for LOF for the mammography dataset with various $\varepsilon$}
    \label{fig:MammoLOFLaplace}
\end{figure}

After having quantitatively analyzed the impact of DP on SHAP values, we now examine the visual interpretation and analysis of SHAP summary plots to illustrate how DP noise influences the interpretability of SHAP explanations visually.
Figure \ref{fig:MammoIforestLaplace} and Figure \ref{fig:MammoLOFLaplace} present the summary plots relative to the mammography dataset for the iForest and LOF models respectively\footnote{The summary plots display SHAP values for each feature and data point, indicating their impact on classifying mammograms as normal or abnormal. On the x-axis, SHAP values show a feature's influence on predictions, with positive or negative values indicating a tendency towards an abnormal or normal prediction, respectively. The y-axis ranks features by importance, and point colors signify feature values—red for high and blue for low}.
\vspace{0mm}
Figure \ref{fig:MammoIforestLaplace} shows that the feature \emph{Contrast} emerges as the most critical feature for the model's prediction, while a \emph{low\_order\_moment} and \emph{Root\_mean\_square\_noise} remain the two least important regardless of the level of privacy $\varepsilon$ introduced. This fact indicates that, for some features, the level of importance remains consistent even after applying DP, regardless of the value of $\epsilon$. Instead, concerning other features, such as \emph{Gradient\_Strength, Area\_of\_object, and average\_gray\_level}, their importance according to SHAP varies with $\epsilon$. For example, for a low level of privacy protection (e.g., $\epsilon = 5$), the consistency concerning the scenario without DP is maintained considering both feature importance and feature contribution, as observable by the similar color distributions in the various scenarios. If, instead, stronger privacy guarantees are set in place (e.g., $\epsilon = 0.01$), both the order and contribution of the three features significantly change. Indeed, the clear distinction between blue and red fades, indicating that the significant level of noise obscures the clarity of SHAP values, rendering the interpretation of output trends more difficult \footnote{As similar trends in the SHAP summary plots are observed in the two other datasets, we omit to show them and the relative discussion. To illustrate the visual changes, we show summary plots for only two $\varepsilon$ values (0.01 and 5).}
Figure \ref{fig:MammoLOFLaplace} shows the summary plots obtained for the LOF model. We observe that the \emph{contrast} feature is consistently the most influential, irrespective of the value $\epsilon$-DP. Specifically, we note that higher values of the contrast feature continue to be highlighted in blue and have a positive impact on the output, indicating their significance, while lower values are consistently highlighted in red and have negative SHAP values. Instead, for $\varepsilon = 0.01$, the SHAP values are no longer distinguishable by color, signifying a less clear impact of feature values on the output of the model. Concerning the other features, there is a variation in their order of influence between the various $\varepsilon$, yet the underlying rationale behind their values remains consistent across most $\varepsilon$ values. However, for a stricter privacy budget of $\varepsilon = 0.01$, there is a notable departure from this consistency, as the distinction between blue and red becomes less clear, thus reducing considerably the interpretability of the model through this plot.
\vspace{0mm}
\vspace{0mm}
These findings align with the ShapGap and ShapLength measures, demonstrating that iForest exhibits a greater sensitivity to DP perturbations compared to LOF in terms of SHAP interpretability. This is evidenced by the more pronounced shift observed in the SHAP summary plots for iForest with decreasing privacy budgets. While both models experience a decline in interpretability for stricter privacy constraints, LOF retains a relatively clearer distinction between influential and non-influential features even at lower $\varepsilon$ values, specifically at $\varepsilon$ equal to 0.01, where iForest's interpretability obscures significantly. Another finding is that while applying DP safeguards individual data privacy through noise injection, this mechanism hindered the interpretability of SHAP summary plots. DP's impact manifests in different ways such as distorted features importance and misleading interpretations. Firstly, the introduced randomness leads to fluctuations in SHAP feature attributions, making it difficult to accurately detect their true impact on model predictions. Secondly, the noise obscured data patterns, diminishing the overall precision of the AD and SHAP values and affecting the extraction of meaningful insights.

\section{Conclusion}\label{Conclusion}
In this paper, we investigate the impact of differential privacy (DP) on the performance and explainability of anomaly detection (AD) models. We compare the performance of Isolation Forest (iForest) and Local Outlier Factor (LOF) under various DP noise conditions and across multiple datasets. The results show that while iForest initially outperforms LOF without DP, LOF exhibits greater robustness to DP. Furthermore, we analyze the impact of DP on explainability by comparing them across different distance metrics, both with and without DP applied. For explainability we use SHapley Additive exPlanations (SHAP). We observe a correlation between the DP parameter ($\varepsilon$) and the magnitude and direction of changes in SHAP values across the metrics. Notably, the impact of DP on SHAP values manifested diversely across datasets and with the different AD techniques. This implies that distinctive data characteristics might affect the sensitivity of SHAP values to DP noise. These findings underscore the trade-off between privacy and explainability when employing DP alongside SHAP values in AD. For future work, we aim to explore techniques to mitigate the effect of DP on SHAP values while upholding adequate privacy guarantees. Additionally, we aim to evaluate the effects of DP on other explainability techniques utilized with deep learning-based AD methodologies.

\bibliographystyle{plain}
{\footnotesize \bibliography{references}}
\end{document}